\documentclass[11pt]{article}
\usepackage{acl}          
\usepackage{latexsym}
\usepackage{microtype}
\usepackage{inconsolata}
\usepackage{graphicx}
\usepackage{booktabs}
\usepackage{multirow}
\usepackage{amsmath}
\usepackage[normalem]{ulem}   

\usepackage{iftex}
\ifPDFTeX
  \usepackage[T1]{fontenc}
  \usepackage[utf8]{inputenc}
  \usepackage{times}
  \newcommand{\ethi}[1]{#1}
  \newcommand{\afr}[1]{#1}
\else
  \usepackage{fontspec}
  \newfontfamily\ethiopicfont{NotoSansEthiopic-Regular.ttf}[
    Path        = ./,
    BoldFont    = NotoSansEthiopic-Bold.ttf]
  \newcommand{\ethi}[1]{{\ethiopicfont #1}}
  \newcommand{\afr}[1]{#1}               
\fi

\title{AfriSwitch: A Benchmark for In-the-Wild African Code-Switched Speech Recognition}
\author{
  Gabrial Zencha Ashungafac\textsuperscript{1} \quad
  Busayo Awobade\textsuperscript{1} \quad
  Tobi Olatunji\textsuperscript{1} \\
  \textsuperscript{1}Intron Health \\
  \texttt{research@intron.io}, \texttt{tobi@intron.io} \\}
\begin{document}
\maketitle

\begin{abstract}
Code-switching is pervasive in bilingual African conversation, yet most ASR
systems assume monolingual input and are evaluated on curated monolingual
benchmarks. We present \textbf{AfriSwitch}, a 61.36-hour human-transcribed
benchmark of in-the-wild code-switched speech spanning 16 African languages
and language varieties, released with switch-level English span tags,
per-utterance Code-Mixing Index (CMI), and switch-point counts. Corpus
statistics show that mixing behaviour varies widely across African languages
along two largely independent axes: how often speakers alternate, and how
balanced the mixture is. No single scalar captures how code-switched a
language is. Benchmarking five open and commercial multilingual ASR systems
zero-shot yields word error rates far above published monolingual figures for
the same languages, with the best system averaging 35.93\% WER and no system
falling below 24\% on any language. Africa-targeted training, not model scale
or nominal language coverage, best predicts performance.
\end{abstract}

\section{Introduction}

Everyday multilingual African communication is rarely monolingual. Speakers
alternate fluidly between English or French, regional lingua francas, and
local languages, often several times within a single utterance. Yet the
benchmarks used to measure African ASR progress assume linguistic consistency
within an utterance, so the systems they rank are optimised for a register
their users do not speak. The result is a measurement gap: reported accuracy
on curated monolingual test sets does not predict how a system behaves in
deployment.

Closing that gap requires an evaluation resource with three properties that no
existing corpus combines: speech that is \emph{naturally} code-switched rather
than synthesised or scripted, coverage broad enough to show how mixing differs
across African languages, and annotation at the level of the switch itself, so
that recognition of embedded-language material can be measured directly
instead of being averaged away into a single error rate.

This paper contributes such a resource and establishes where current systems
stand on it. We release 61.36 hours of human-transcribed conversational
code-switched speech across 16 African languages and language varieties, with
per-utterance mixing statistics and switch-level English span tags, and we
benchmark five open and commercial multilingual ASR systems on it. Every
system degrades sharply relative to its published monolingual numbers, and the
degradation is not explained by model scale or by how many languages a system
nominally supports.

\paragraph{Contributions.}
\begin{itemize}\setlength\itemsep{0pt}
  \item \textbf{AfriSwitch benchmark}: 61.36 hours of human-transcribed conversational code-switched speech across 16 African languages and language varieties, with switch-level English span tags, per-utterance CMI, and switch-point counts.
  \item \textbf{Code-switching characterisation}: CMI and $S^{*}$ statistics documenting substantial and previously unreported variation in mixing behaviour across African languages.
  \item \textbf{Zero-shot evaluation}: Five open and commercial multilingual ASR systems benchmarked on natural code-switched speech, revealing large WER gaps relative to monolingual benchmarks for the same languages, scored with per-language normalizers written by linguists and validated by native-speaker annotators.
  \item \textbf{Qualitative error analysis}: Three recurring failure modes grounded in system transcripts (\S\ref{sec:qualitative}), including deletion of embedded spans that aggregate WER cannot see, and script-nativisation of embedded names scored as error regardless of whether the words were recognised.
\end{itemize}

\section{Related Work}

\paragraph{African speech resources and benchmarks.}
Publicly available speech data for African languages has grown rapidly,
including AfriSpeech-200 \cite{olatunji-etal-2023-afrispeech}, Google Waxal
\cite{waxal2026}, NCHLT \cite{Barnard2014TheNS}, ALFFA \cite{gelas,Abate2005},
Mozilla Common Voice \cite{ardila-etal-2020-common}, Africa Next Voices
\cite{za-african-next-voices-2025,digitalumuganda2025afrivoice_kinyarwanda},
and Naija Voices \cite{emezue2025}. Evaluation suites such as AfriVox
\cite{awobade2025afrivox}, FLEURS \cite{fleurs2022arxiv}, SimbaBench
\cite{elmadany-etal-2025-voice}, and AfriSpeech-MultiBench
\cite{ashungafac-etal-2025-afrispeech} now support systematic assessment across
languages and domains. Afrispeech-Dialog \cite{sanni-etal-2025-afrispeech}
moves closer to deployment conditions by evaluating spontaneous African-accented
English conversation, reporting over 10\% degradation relative to native
accents. All of these, however, score utterances assumed to be in one language.

\paragraph{Code-switched speech corpora.}
The reference corpora for code-switched ASR come almost entirely from outside
Africa. SEAME \cite{lyu-etal-2010-seame} established the template with
Mandarin--English conversational speech from South-East Asia, and the ASRU 2019
challenge \cite{shi2020asru} built open training and evaluation sets on the same
pair. For African languages the principal resource remains the South African
soap opera corpus \cite{vanderwesthuizen-niesler-2018-first,
yılmaz2018buildingunifiedcodeswitchingasr}, 14.3 hours across four
English--Bantu pairs; because it is drawn from scripted broadcast dialogue, it
captures written-then-performed switching rather than spontaneous conversational
switching. CS-FLEURS \cite{yan2025csfleursmassivelymultilingualcodeswitched}
achieves broad language coverage by synthesising code-switched utterances from
read monolingual speech, and SwitchLingua \cite{xie2025switchlingua} scales
generated code-switched text and audio to 12 languages. Synthetic construction
controls the mixing distribution but reproduces neither the acoustic conditions
nor the pragmatics of real bilingual talk, and none of these resources carries
switch-level language tags on natural speech.

\paragraph{Benchmarks for code-switched language technology.}
LinCE \cite{aguilar-etal-2020-lince} and GLUECoS
\cite{khanuja-etal-2020-gluecos} centralised evaluation for code-switched NLP
and demonstrated the value of a shared benchmark, but both are text-only and
neither covers an African language pair.

\paragraph{Modelling and evaluating code-switched ASR.}
Massively multilingual pre-training has improved cross-language generalization
\cite{radford2022robustspeechrecognitionlargescale,pratap2023scalingspeechtechnology1000,omnilingualasrteam2025},
but these models are trained and evaluated on utterances assumed monolingual.
Proposed remedies include synthetic data mixing
\cite{babatunde-etal-2025-beyond,dhawan2023unified,pratapa2018languagemodeling},
language prompt fusion \cite{yang2023adapting}, attention-guided adaptation
\cite{aditya2024attention}, and encoder refining with language-aware decoding
\cite{zhao2025adapting}. Closest to our setting,
\citet{ogunremi-etal-2023-multilingual} fine-tune self-supervised multilingual
representations for four code-switched South African languages, cutting WER by
up to 20 points absolute and showing that transfer from multilingual
pre-training outperforms training from scratch when code-switched data is
scarce. Surveying 127 end-to-end code-switching ASR papers,
\citet{agro2025codeswitching} find that a small subset of language pairs
absorbs most of the attention and that evaluation practice is inconsistent
across the literature. On the metric side, \citet{ugan2025pier} show that
aggregate WER systematically under-weights embedded-language tokens and propose
restricting scoring to points of interest, an analysis that requires
switch-level annotation of the kind AfriSwitch provides. Our work is
complementary to this modelling literature: we supply the natural, broadly
multilingual, switch-annotated evaluation data against which such methods can
be tested.

\section{AfriSwitch Dataset}

\subsection{Language Selection and Data Sources}

To span multiple African regions and both major colonial-language contact settings, we cover \textbf{West Africa} (Hausa, Igbo, Yoruba, Nigerian Pidgin, Akan, Wolof), \textbf{East Africa} (Swahili, Kinyarwanda, Amharic, Oromo, Luganda), \textbf{Southern Africa} (Zulu, Shona, Tswana, Afrikaans), and \textbf{francophone} contexts (French). Following \citet{li2023yodas}, audio was drawn from publicly available YouTube videos and podcasts under permissive licenses, selected by bilingual annotators on African crowdsourcing platforms \cite{olatunji-etal-2023-afrispeech} for the presence of code-switching. The dataset is publicly available under a Creative Commons Attribution 4.0 International (CC BY 4.0) license.\footnote{\url{https://huggingface.co/datasets/intronhealth/AfriSwitch}}

\subsection{Processing and Annotation}
\label{sec:annotation}

Audio was segmented using VAD; shorter segments were concatenated up to 40 seconds to increase the probability of capturing code-switches, following \citet{koluguri2025granary}. Bilingual native speakers (college-educated, ages 18--35, paid \$10--\$50/hour) produced verbatim transcriptions. Transcription quality was controlled in two stages: initial transcription and review, followed by independent meta-review by a second annotator.

\paragraph{Token-level switch tagging.}
Embedded English is marked at the \textbf{token level}: each token in a
transcript carries a binary label (English / non-English), and maximal runs of
English tokens form the spans released in the parallel
\texttt{transcription\_tagged} field using \texttt{[[EN]]}\,\dots\,\texttt{[[/EN]]}
delimiters. Tagging is automatic, validated against human span annotations on a
subset of clips in each language. The per-utterance CMI and switch-point counts
in Table~\ref{tab:dataset_stats} are derived from these labels.

For the present release, two additional processing steps were applied. First, utterances were \textbf{character-rate filtered}: those falling outside each language's 5th--95th percentile of characters-per-second were removed as likely audio/text misalignments. Second, \textbf{code-mixing metrics} (per-utterance CMI and switch-point count) were computed for every retained utterance.

\subsection{Dataset Structure}

Each example contains: \texttt{audio} (16\,kHz HuggingFace \texttt{Audio} feature), \texttt{language} (primary/matrix language of the utterance), \texttt{filename}, \texttt{transcription} (verbatim plain text), \texttt{transcription\_tagged} (English spans delimited as above), \texttt{cmi}, \texttt{num\_switch\_points}, and \texttt{duration}. Each language is released as a separate configuration with a single \texttt{test} split.

\subsection{Dataset Statistics}

Table~\ref{tab:dataset_stats} summarizes the corpus: 61.36 hours across 18,861 utterances containing 78,333 code-switch events. To characterize the degree of mixing, we report two complementary metrics for each language variety.

\paragraph{Switch points ($S^{*}$).}
Following \citet{gambäck-das-2016}, a switch point is defined as any token preceded by a token with a different language tag. We count the number of such alternation points per utterance and report both the per-utterance average and the corpus total.

\paragraph{Code-Mixing Index (CMI).}
We adopt the CMI formulation of \citet{das-gambäck-2014}:
\begin{equation}
  \mathrm{CMI} =
  \begin{cases}
    100 \times \left[1 - \dfrac{\max(w_i)}{n - u}\right] & n > u \\[6pt]
    0 & n = u
  \end{cases}
  \label{eq:cmi}
\end{equation}
where $w_i$ is the token count for language $i$, $\max\{w_i\}$ is the count for the most prominent language, $n$ is the total number of tokens, and $u$ is the number of language-independent tokens. A CMI of~0 indicates a monolingual utterance; higher values reflect a more balanced distribution across languages \citep{srivastava-singh-2021}.

\subsection{All Code-Switching is NOT Equal}

Corpus-level CMI ranges from 4.19 (Nigerian Pidgin) to 28.20 (Akan), and average $S^{*}$ from 1.46 (Pidgin) to 10.29 (Swahili), reflecting genuine differences in mixing density, in the average number of language alternations per utterance, and in annotation granularity across languages.

The two metrics capture distinct properties and do not move together. Swahili has by far the highest alternation rate (10.29 switches per utterance) but a CMI of 25.72, comparable to Luganda (25.64) at less than half the switch rate: Swahili speakers in our data alternate frequently but in short embedded bursts, whereas Luganda utterances alternate less often but in more balanced blocks. Akan is the clearest case of the two axes pulling apart, pairing the highest CMI in the corpus (28.20) with one of the lowest switch rates (2.89): its utterances alternate rarely, but when they do, the two languages are contributed in near-equal measure. Nigerian Pidgin sits at the opposite extreme on both measures (1.46, 4.19), consistent with its status as a stabilised contact variety in which English-origin material is lexicalised rather than switched into. Amharic and Hausa likewise show near-average switch rates (4.60, 4.00) with low CMI (13.11, 13.09), indicating utterances that remain firmly anchored in the matrix language despite regular English insertions. Any single scalar summary of ``how code-switched'' a language is would collapse these distinctions, and we therefore release both metrics per utterance.

\begin{table}[t]
  \centering
  \footnotesize
  \caption{AfriSwitch corpus statistics. \textbf{Avg $S^{*}$}~=~average number of switch points per utterance \citep{gambäck-das-2016}. \textbf{CMI}~=~Code-Mixing Index \citep{das-gambäck-2014}.}
  \label{tab:dataset_stats}
  \setlength{\tabcolsep}{2pt}
  \begin{tabular}{lrrrrr}
    \toprule
    \textbf{Lang} & \textbf{Hrs} & \textbf{Utt} &
    \textbf{Avg $S^{*}$} & \textbf{Total $S^{*}$} & \textbf{CMI} \\
    \midrule
    Kinyarwanda   & 5.00 & 1{,}577 &  4.51 &  7{,}108 & 16.95 \\
    Amharic       & 5.00 & 1{,}229 &  4.60 &  5{,}650 & 13.11 \\
    Zulu          & 5.00 & 1{,}465 &  4.40 &  6{,}447 & 24.76 \\
    Igbo          & 5.00 & 1{,}848 &  4.10 &  7{,}575 & 27.64 \\
    Yoruba        & 5.00 & 1{,}877 &  5.33 & 10{,}002 & 22.93 \\
    Hausa         & 5.00 & 1{,}515 &  4.00 &  6{,}053 & 13.09 \\
    Akan          & 5.00 & 1{,}345 &  2.89 &  3{,}881 & 28.20 \\
    Pidgin        & 4.56 & 1{,}801 &  1.46 &  2{,}621 &  4.19 \\
    Oromo         & 4.25 & 1{,}217 &  2.95 &  3{,}586 & 13.39 \\
    Swahili       & 3.89 &   650   & 10.29 &  6{,}689 & 25.72 \\
    Shona         & 3.86 & 1{,}155 &  4.85 &  5{,}599 & 24.55 \\
    French        & 3.22 &   903   &  3.00 &  2{,}713 & 10.92 \\
    Tswana        & 2.74 &   805   &  4.38 &  3{,}529 & 23.22 \\
    Wolof         & 1.95 &   914   &  4.95 &  4{,}524 & 17.48 \\
    Luganda       & 1.21 &   362   &  3.97 &  1{,}437 & 25.64 \\
    Afrikaans     & 0.68 &   198   &  4.64 &    919   & 13.43 \\
    \midrule
    \textbf{Total} & \textbf{61.36} & \textbf{18{,}861} &
    \textbf{4.15} & \textbf{78{,}333} & \textbf{18.90} \\
    \bottomrule
  \end{tabular}
\end{table}

\section{Benchmarking Multilingual ASR Systems}

\subsection{Evaluated Models and Protocol}
\label{sec:protocol}

We evaluate five multilingual ASR systems spanning open research models, production APIs, and Africa-focused systems: \textbf{Sahara V2} and \textbf{Sahara V2.5} \cite{doc:sahara} (Africa-optimized production systems), \textbf{Omnilingual LLM 7B} \cite{omnilingualasrteam2025} (1{,}600+ language coverage), \textbf{Gemini 3.6} \cite{doc:gemini} (multimodal, commercial), and \textbf{ElevenLabs} \cite{doc:elevenlabs} (commercial speech-to-text API).

All models are evaluated zero-shot, with no fine-tuning on AfriSwitch or on any code-switched data. Switch-level language tags are removed from references prior to scoring.

\paragraph{Language conditioning.}
Every system is given the dominant (matrix) language of each utterance, supplied through whichever interface that system documents: a language prompt for Gemini~3.6, the dominant-language identifier for Omnilingual~LLM~7B as its usage guidelines require, and the equivalent language specification for Sahara~V2, Sahara~V2.5, and ElevenLabs. Conditioning is therefore matched across the comparison, and no system has to infer the matrix language before transcribing.

This is deliberately an \emph{optimistic} setting for code-switched input: naming the matrix language resolves half of the language-identification problem for free. Real deployments rarely have per-utterance language labels, so the figures below should be read as an upper bound on what these systems would achieve unprompted.

\subsection{Text Normalization and Scoring}
\label{sec:normalization}

Reported WER is sensitive to normalization choices, and code-switched speech makes those choices more consequential than usual: the same pipeline must handle two orthographies at once. We therefore specify the procedure in full.

Evaluation involves a single processing stage. Model output and reference are passed through the same pre-processing pipeline immediately before WER is computed, and nothing is applied to either side afterwards. Every rule below is therefore applied \emph{identically and symmetrically} to hypothesis and reference; no rule is applied to one side only. In particular, system output receives no correction or repair of any kind: our pipeline supports an optional LLM-based transcript correction pass, but enabling it would confound the comparison, since the correcting model's own coverage of each language would become part of what is measured.

\paragraph{Pre-processing rules.} In order:
\begin{enumerate}\setlength\itemsep{0pt}
  \item \textbf{Annotation-marker removal.} A closed list of transcription markers is deleted case-insensitively from both sides, covering inaudible, silence, music, and noise tags in bracket, parenthesis, and bare forms, together with their attested misspellings.
  \item \textbf{Filler-word removal.} A closed list of general fillers (\textit{ah, eh, hmm, huh, mm, mmhmm, oh, uh, uhhuh, um, umhum}, and variants) is removed by exact token match, along with inline hesitation forms and ellipses. Fillers are transcribed inconsistently by human annotators and emitted inconsistently by ASR systems, so scoring them measures annotation convention rather than recognition.
  \item \textbf{Numeric canonicalization.} Spelled-out cardinals and ordinals are converted to digits on both sides, so that \textit{twenty five} and \textit{25}, or \textit{third} and \textit{3rd}, are not counted as substitutions. Systems differ arbitrarily in whether they emit numbers as words or digits, as do human transcribers; without this step the metric would penalise a formatting convention rather than a recognition error.
  \item \textbf{Verbalized punctuation.} Dictated punctuation (\textit{comma}, \textit{full stop}, and the localized variants \textit{koma}, \textit{coma}) is mapped to its orthographic form before punctuation is stripped.
  \item \textbf{Typographic normalization.} Curly quotes, apostrophes, backticks, hyphens, brackets, colons, semicolons, and terminal punctuation are replaced by whitespace, so that hyphenated and spaced forms of the same compound do not differ.
  \item \textbf{Case folding and whitespace.} Text is lowercased; tabs and newlines become spaces; runs of whitespace are collapsed; leading and trailing whitespace is stripped.
\end{enumerate}
These rules are language-independent and follow a Whisper-style basic text normalizer \cite{radford2022robustspeechrecognitionlargescale}, with the English normalizer reserved for the monolingual English condition.

\paragraph{Language-specific normalization.}
Language-independent rules are not sufficient here. Orthographic conventions differ across the corpus in ways that a generic normalizer either ignores or actively damages, and unresolved variation is charged as recognition error even when the system produced an acceptable transcript. We therefore developed a dedicated normalizer for each of the 12 benchmarked languages. Rules were written by linguists with competence in the language and then validated by contributors drawn from the same pool that annotated the data, so that each rule was checked by native speakers who had seen how the convention behaves in real transcripts rather than only in the abstract. Rules address, among other categories, accepted variant spellings of the same lexeme, word-boundary and clitic-attachment conventions, and the nativized spellings under which English loanwords are conventionally written in each orthography. The last of these is especially consequential in a code-switched corpus, where the same borrowed word may appear in either its English or its nativized form within a single utterance. As with the shared rules, each language's normalizer is applied identically to hypothesis and reference. The rule sets themselves are not part of the public release.

\paragraph{Diacritics are preserved.}
The normalizer can optionally strip combining diacritics. \textbf{We do not use this option: every language whose orthography carries diacritics is scored with diacritics intact.} Stripping them is common practice and inflates apparent accuracy, but it does so by introducing lexical ambiguity that is not present in the language. In Yoruba, the unmarked string \textit{igba} corresponds to at least four distinct lexemes once tone and vowel quality are written (`two hundred', `calabash', `time', `garden egg'), and \textit{oko} likewise spans `farm', `husband', `vehicle', and `spear'. In Igbo, \textit{akwa} realises `cry', `bed', `cloth', and `egg' depending on tone. Under diacritic-stripped scoring, a system that emits the bare consonant--vowel skeleton is credited with recognising a word it never disambiguated, and the metric silently rewards models that ignore the tonal layer of the language. Since AfriSwitch includes tone-marked orthographies (Yoruba and Igbo most consequentially, alongside Wolof, French, Afrikaans, and Hausa), diacritic-preserving WER is the only figure that reflects what was actually said. All numbers in Table~\ref{tab:baseline_results} are diacritic-preserving; the diacritic-stripped variant is computed by our pipeline but is not reported.

\paragraph{Scoring.}
WER is computed at the corpus level with \texttt{jiwer}: total edit operations divided by total reference words, rather than the mean of per-utterance rates, so that long utterances are not down-weighted relative to short ones. If either side is empty after normalization it is replaced by a sentinel token, ensuring that an empty hypothesis is charged as a full deletion error rather than skipped. Per-utterance WER and CER are retained alongside the corpus figures to support the error analysis in \S\ref{sec:discussion}.

\subsection{Results}
\label{sec:results}

Table~\ref{tab:baseline_results} reports zero-shot WER. Every system performs substantially worse than published figures for the same languages on monolingual benchmarks \cite{awobade2025afrivox,elmadany-etal-2025-voice}, confirming that naturally occurring conversational code-switching is a significant failure mode rather than an incremental difficulty.

\paragraph{Africa-targeted training dominates.}
Sahara V2.5 achieves the lowest WER on 10 of the 12 benchmarked languages and the lowest average WER overall (35.93\%), ahead of Omnilingual LLM 7B (51.46\%), Gemini 3.6 (55.05\%), ElevenLabs (56.48\%), and Sahara V2 (59.90\%). The margin over the general-purpose commercial systems is large and consistent: on Amharic, Sahara V2.5 reaches 24.58\% against 45.11\% for Gemini 3.6 and 51.80\% for ElevenLabs. The V2\,$\rightarrow$\,V2.5 improvement within a single system family (59.90\%\,$\rightarrow$\,35.93\% average) is itself the largest single effect in the table, and exceeds the spread across all other systems, indicating that targeted data and training decisions matter more here than raw scale or breadth of language coverage.

\paragraph{Coverage is not competence.}
Omnilingual LLM 7B nominally supports over 1{,}600 languages, yet is outperformed on 11 of 12 languages by a far smaller Africa-targeted system, and exceeds 75\% WER on Igbo (75.52\%) and 85\% on Luganda (85.33\%). Nominal support for a language in a massively multilingual model is therefore a weak predictor of usable accuracy on conversational code-switched speech in that language.

\paragraph{Some languages are hard for everyone.}
Luganda (81.17\% mean WER across systems), Yoruba (71.47\%), and Igbo (69.46\%) remain difficult for every system evaluated, while Afrikaans (35.79\%), Pidgin (38.38\%), and Swahili (38.42\%) are comparatively tractable. Luganda is the hardest language for four of the five systems, with only Sahara V2.5 bringing it below 50\% WER. This ordering does not track corpus size in AfriSwitch, and we take it to reflect the amount and quality of each language's representation in the systems' training data rather than any property of the benchmark itself.

\begin{table}[t]
\centering
\footnotesize
\caption{Zero-shot WER (\%) on AfriSwitch, over the 12 languages benchmarked to date. Best per language in \textbf{bold}. Sahara V2.5 is best on 10 of 12.}
\label{tab:baseline_results}
\setlength{\tabcolsep}{2.5pt}
\begin{tabular}{lccccc}
\toprule
& \multicolumn{2}{c}{\textbf{Sahara}} & \textbf{Omni.} & \textbf{Gemini} & \textbf{Eleven} \\
\cmidrule(lr){2-3}
\textbf{Lang} & \textbf{V2} & \textbf{V2.5} & \textbf{7B} & \textbf{3.6} & \textbf{Labs} \\
\midrule
Kinyarwanda & 56.34 & \textbf{30.73} & 60.41 & 40.74 & 46.58 \\
Amharic     & 50.35 & \textbf{24.58} & 51.99 & 45.11 & 51.80 \\
Zulu        & 81.01 & 49.30 & \textbf{48.19} & 50.76 & 62.02 \\
Igbo        & 83.58 & \textbf{46.94} & 75.52 & 75.88 & 65.37 \\
Yoruba      & 89.66 & \textbf{45.54} & 70.27 & 77.26 & 74.62 \\
Hausa       & 45.56 & \textbf{31.77} & 39.27 & 45.44 & 41.01 \\
Pidgin      & 42.19 & \textbf{32.83} & 42.71 & 38.84 & 35.33 \\
Swahili     & 43.27 & \textbf{34.12} & 39.22 & 36.44 & 39.03 \\
Luganda     & 89.95 & \textbf{43.87} & 85.33 & 95.97 & 90.73 \\
Afrikaans   & 59.14 & 26.45 & 30.43 & 36.90 & \textbf{26.04} \\
Akan        & \textbf{34.52} & 36.67 & 37.41 & 52.80 & 78.83 \\
Wolof       & 43.18 & \textbf{28.33} & 36.79 & 64.42 & 66.39 \\
\midrule
\textbf{Avg} & 59.90 & \textbf{35.93} & 51.46 & 55.05 & 56.48 \\
\bottomrule
\end{tabular}
\end{table}

\subsection{Discussion}
\label{sec:discussion}

Two observations cut across the results. First, the gap between the best and worst system on a given language is frequently larger than 40 WER points (Luganda: 43.87 vs.\ 95.97; Yoruba: 45.54 vs.\ 89.66), meaning that system choice, rather than the intrinsic difficulty of the language, currently dominates outcomes for African code-switched speech. Second, even the strongest system leaves roughly a quarter to a half of words wrong across the benchmark. At these error rates, downstream applications that depend on faithful transcription of bilingual conversation are not yet supportable in most of these languages.

We also note a limitation of WER as the sole metric in this setting. Because the matrix language supplies the majority of tokens in most utterances, aggregate WER is dominated by matrix-language performance, and systematic failures on embedded English words, deletion at switch boundaries in particular, can be substantially masked \citep{ugan2025pier}. AfriSwitch releases switch-level English span tags precisely so that point-of-interest metrics such as PIER \citep{ugan2025pier} can be computed on natural conversational speech, which existing code-switched benchmarks do not permit. We leave such an analysis to future work.

\subsection{Qualitative Analysis}
\label{sec:qualitative}

Table~\ref{tab:baseline_results} reports error rates in aggregate, but a
single WER number cannot distinguish a system that mistranscribes a
difficult word from one that silently deletes an entire embedded-language
span, nor does it explain \emph{why} a given language is hard. We manually
inspected outputs from the four systems for which we retain raw
transcripts (Sahara~V2.5, Omnilingual~LLM~7B, Gemini~3.6, and ElevenLabs)
and surface three recurring failure patterns below, each grounded in a
representative utterance. Reference transcriptions and system hypotheses
are reproduced verbatim except for minor re-wrapping for column width; the
span most relevant to each discussion is underlined.

\paragraph{Switch-boundary deletion is invisible to aggregate WER.}
In the Swahili utterance below, Omnilingual~LLM~7B produces a fluent,
grammatical monolingual Swahili sentence that simply omits the embedded
English noun phrase \emph{plaster work} at all three points where it
occurs in the reference, rather than mishearing it.

\begin{quote}\small
\textbf{Reference:} \afr{Tunafanya kazi ya \uline{plasta work} na leo
tumefanya extenary \uline{plasta work}. Tumeshamaliza
\uline{plaster weki} kwa asilimia 90 na wiki hii tunatarajia kuanza
kufanya \uline{plasta} lakini kwa kuta za ndani.}\\[2pt]
\textbf{Sahara V2.5:} \afr{Tunafanya kazi ya \uline{plasta work}
ambayo tumefanya extenary \uline{plasta work}. Tumeshamaliza
\uline{plasta work} kwa asilimia 90 na wiki hii tunatarajia kuanza
kufanya \uline{plasta work} kwa kuta za ndani.}\\[2pt]
\textbf{Omnilingual 7B:} \afr{Tunafanya kazi ya ambayo tumefanya
tumeshamaliza kwa asilimia 90 na wiki hii tunatarajia kuanza kufanya kwa
kuta za ndani.}\\[2pt]
\textbf{Gemini 3.6:} \afr{Tunafanya kazi ya \uline{plaster work}
ambayo tumefanya external \uline{plaster work}. Tumeshamaliza
\uline{plaster work} kwa asilimia 90 na wiki hii tunatarajia kuanza
kufanya \uline{plaster work} kwa kuta za ndani.}\\[2pt]
\textbf{ElevenLabs:} \afr{Tunafanya kazi ya \uline{plaster work}
ambayo tumefanya external \uline{plaster work}. Tumeshaamaliza
\uline{plaster work} kwa asilimia tisini na wiki hii tunatarajia
kuanza kufanya \uline{plaster work} kwa kuta za ndani.}
\end{quote}

Because the deleted phrase contributes only a handful of the utterance's
33 reference words, this triple deletion moves corpus-level WER by well
under the gap separating Omnilingual (39.22\%) from Sahara~V2.5 (34.12\%)
on Swahili in Table~\ref{tab:baseline_results}, while eliminating
essentially all of the code-switched content the utterance contains. This
is precisely the failure mode point-of-interest metrics such as PIER
\citep{ugan2025pier} are designed to surface and that aggregate WER,
dominated by matrix-language tokens, structurally cannot
(\S\ref{sec:discussion}).

\paragraph{Script separability shapes what counts as a switch.}
Amharic's Ethiopic script makes English material visually separable by
character set. Systems disagree on whether to preserve that separation.

\begin{quote}\small
\textbf{Reference:} \uline{professor}\ethi{ አንድሪያስ እስከ ሁለተኛ ደረጃ
ትምህርታቸው ድረስ እዚሁ አዲስ አበባ የተማሩ ሲሆን ከዚያ በኋላም ወደ }\uline{America}\ethi{
በማቅናት በ }\uline{Williams College}\ethi{ እና በ }\uline{Yale University}\ethi{
ፍልስፍናን ተምረው ዶክትሬታቸውን አግኝተዋል።}\\[2pt]
\textbf{Sahara V2.5:} \uline{professor}\ethi{ አንድርያስ እስከ }2\ethi{ኛ ደረጃ
ትምህርታቸው ድረስ እዚሁ አዲስ አበባ የተማሩ ሲሆን ከዚያ በኋላም ወደ }\uline{America}\ethi{
በማቅናት በ }\uline{Williams college}\ethi{ እና በየል ዩኒቨርሲቲ ፍልስፍናን ተምረው
ዶክትሬታቸውን አግኝተዋል።}\\[2pt]
\textbf{Omnilingual 7B:} \ethi{ፕሮፌሰር አንድሪያስ እስከ ሁለተኛ ደረጃ ትምህርታቸው ድረስ
እዚሁ አዲስ አበባ የ ተማሩ ሲሆን ከዚያ በኋላ ም ወደ አሜሪካ በ ማቅናት በ }\uline{\ethi{ዊልያምስ
ኮሌጅ}}\ethi{ ና በ የ ል ዩኒቨርሲቲ ፍልስፍና ን ተምረው ዶክትሬታቸው ን አግኝተ ዋል}\\[2pt]
\textbf{Gemini 3.6:} \ethi{ፕሮፌሰር አንድሪያስ ሁለተኛ ደረጃ ትምህርታቸውን እዚሁ አዲስ
አበባ የተማሩ ሲሆን ከዚያ በኋላም ወደ አሜሪካ በመጓዝ በ}\uline{\ethi{ዊሊያምስ ኮሌጅ}}\ethi{ እና
በ}\uline{\ethi{ዬል ዩኒቨርሲቲ}}\ethi{ ፍልስፍናን ተምረው ዶክተሬታቸውን አግኝተዋል።}\\[2pt]
\textbf{ElevenLabs:} \ethi{ፕሮፌሰር አንድሬያስ እስከ ሁለተኛ ደረጃ ትምህርታቸው ድረስ
እዚህ አዲስ አበባ የተማሩ ሲሆን ከዚህ በኋላም ወደ አሜሪካ በማቅናት በ}\uline{William
College}\ethi{ እና በ}\uline{Yale University}\ethi{ ፍልስፍናን ተምርው ዶክተሬት አጥቶን
አግኝተዋል።}
\end{quote}

The underlying ambiguity is whether \emph{professor} is a switch into
English at all, or an established Amharic loanword conventionally spelled
\ethi{ፕሮፌሰር} (\S\ref{sec:normalization}).
Sahara~V2.5 and ElevenLabs answer ``switch,'' preserving the Latin script
the reference transcriber chose; Omnilingual and Gemini answer
``loanword,'' nativizing every instance to Ethiopic script, including
institution names such as \emph{Williams College} and \emph{Yale
University} that are not conventional loanwords at all, an
overgeneralization scored as error regardless of whether the underlying
words were recognized correctly. Only ElevenLabs preserves both
institution names in Latin script matching the reference, while still
nativizing \emph{professor} itself, showing the choice is made per-token
rather than per-system.

\paragraph{A creole's matrix/embedded boundary defeats every system in
the same place.} Nigerian Pidgin is an English-lexified creole: most of
its vocabulary is English-derived, so the boundary between matrix and
embedded material is not cleanly recoverable at the word level. That
difficulty resurfaces acoustically, concentrated at the ambiguous
boundary tokens rather than spread across the utterance.

\begin{quote}\small
\textbf{Reference:} \uline{gats} continue with the person how long
you inten to do music for now course I dey do am professionally now
How long you wan do am for\\[2pt]
\textbf{Sahara V2.5:} You \uline{gatz} continue with the person.
How long you intend to do music for? Now cause I do am professionally
now. How long you want do am for?\\[2pt]
\textbf{Omnilingual 7B:} \uline{guys} continue with the person how
how long you intend to do music for for now because are you doing
professionally now how long you want doing for\\[2pt]
\textbf{Gemini 3.6:} \uline{Gas} continue with the person. How, how
long you intend to do music for? For now cause I dey do am
professionally now. How long you want do am for?\\[2pt]
\textbf{ElevenLabs:} Hm \uline{You guys} continue with the person I,
how, how long do you intend to do music for? For now, 'cause I dey do am
professionally now How long you want do am for?
\end{quote}

Four systems produce four different tokens for the opening
word (\emph{gatz}, \emph{guys}, \emph{Gas}, or nothing at all), with no
two systems agreeing and none matching the transcriber's \emph{gats}
exactly. Tellingly, every system transcribes \emph{I dey do am
professionally}, an unambiguous Pidgin construction with no English
paraphrase, correctly and near-identically a few words later. The
difficulty concentrates exactly at the boundary token, not at Pidgin
transcription broadly, consistent with Pidgin's comparatively low mean
WER (38.38\%; \S\ref{sec:results}) despite the creole's structural
ambiguity.

\section{Conclusion}

We presented AfriSwitch, a 61.36-hour human-transcribed in-the-wild code-switched benchmark spanning 16 African languages and language varieties, released with per-utterance Code-Mixing Index, switch-point counts, and switch-level English span tags. Corpus statistics show that code-switching behaviour varies widely across African languages along two largely independent axes, alternation frequency and mixing balance, so that no single scalar characterises how code-switched a language is. Zero-shot benchmarking of five multilingual ASR systems shows word error rates far above those reported on monolingual benchmarks for the same languages, with the best system averaging 35.93\% WER and no system falling below 24\% on any language. Africa-targeted training, not model scale or nominal language coverage, is the strongest predictor of performance in our results. Future work will extend evaluation to the remaining language configurations, use the released switch-level tags for point-of-interest analysis of embedded-language recognition, and study adaptation strategies for closing the gap on natural conversational speech.



\section*{Limitations}

\paragraph{Language and geographic coverage.} AfriSwitch spans 16 languages and language varieties across four broad regions, but the continent hosts over 2{,}000 languages. The selected languages skew toward relatively higher-resourced African languages with existing ASR infrastructure. Coverage is also uneven within the corpus: seven languages have the full 5.00 hours, while Afrikaans (0.68 hrs) and Luganda (1.21 hrs) are represented by under 400 utterances each, so per-language WER for these should be read with correspondingly wide error bars.

\paragraph{ASR evaluation metric.} We report WER only. As discussed above, WER under-weights embedded-language tokens in code-switched speech and can mask systematic failures at switch boundaries \citep{ugan2025pier}. The released tags support point-of-interest metrics, but we do not report them here.

\paragraph{Comparability of reported WER.} Because we score with diacritics preserved (\S\ref{sec:normalization}), our figures for Yoruba, Igbo, Wolof, Afrikaans, and Hausa are not directly comparable to published results that strip diacritics before scoring, and will read as higher for the same system output. We regard this as the correct trade-off, since diacritic-stripped WER credits systems for contrasts they did not make, but comparisons against externally reported numbers for these languages should be treated as indicative rather than exact.

\paragraph{Domain and register.} Audio is sourced from YouTube videos and podcasts, which over-represent broadcast, interview, and public-discourse registers relative to private conversation, and skew toward speakers with media presence.

\section*{Ethical Considerations}

\paragraph{Annotator compensation and working conditions.} All transcription annotators were bilingual native speakers recruited through African crowdsourcing platforms and compensated at \$10--\$50 per hour, above local market rates. Annotators were informed of the nature of the task, worked flexible hours, and were not exposed to harmful or sensitive content.

\begin{figure}[ht!]
    \centering
    \includegraphics[width=1\linewidth]{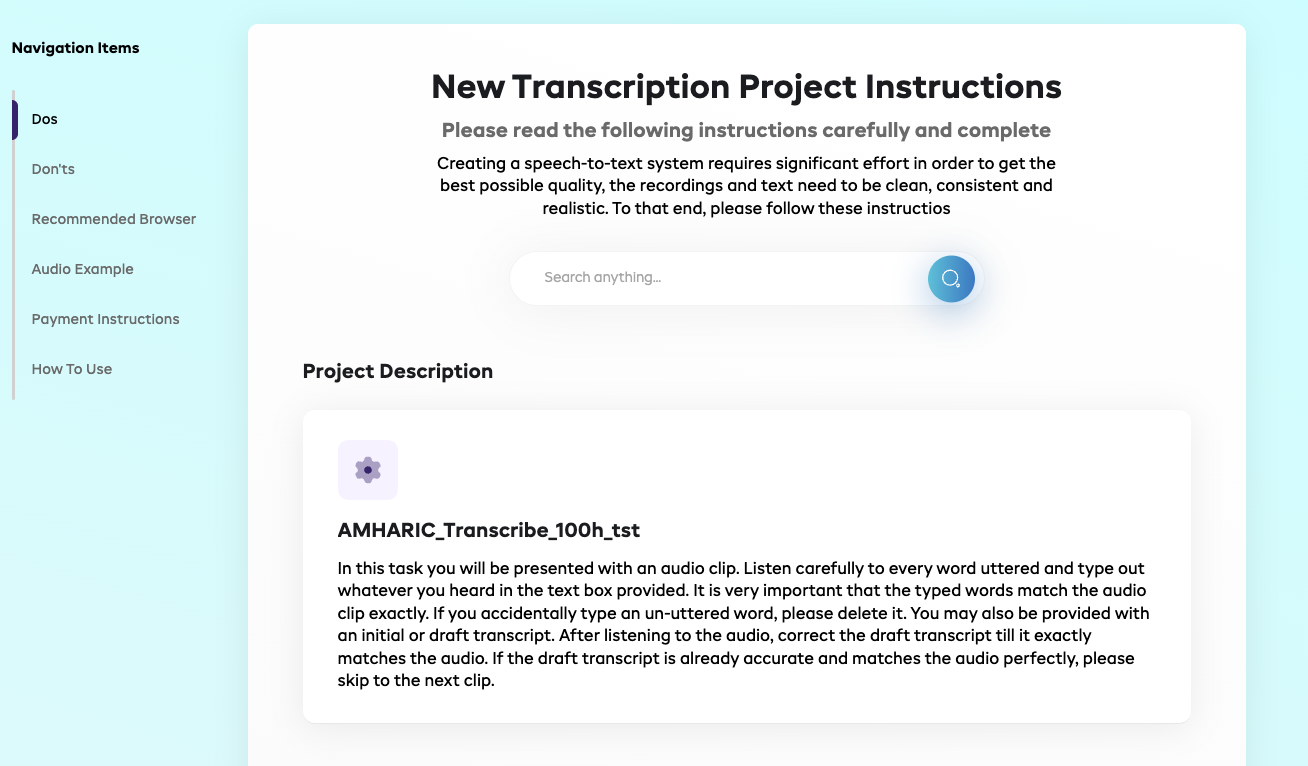}
    \caption{Annotator instructions.}
    \label{fig:annotator_instructions}
\end{figure}

\paragraph{Data provenance.} Audio was sourced from publicly available YouTube videos and podcasts under permissive licenses. Annotators performed verbatim transcription only; no original recordings were made. The released dataset contains transcriptions and processed audio segments only, without links to original sources, preventing direct tracing back to the original content creators.

\paragraph{Annotator biases.} Transcription annotators were college-educated bilingual speakers aged 18--35. While this ensures linguistic competence, it may introduce biases in how disfluencies, informal registers, and non-standard orthography are handled. No annotator demographic information is included in the released benchmark.

\paragraph{Language representation.} By focusing on 16 languages and varieties, this work implicitly prioritizes these communities over the many other African language communities not represented. We encourage the community to extend this benchmark to a broader set of languages, particularly those with fewer existing resources.

\bibliography{afriswitch}

\end{document}